\documentclass{article}

\usepackage{graphicx}   % 插入图片宏包
\usepackage{pdfpages}   % 插入PDF宏包
\usepackage{amsmath}    % 支持数学公式
\usepackage{amssymb}    % 额外数学符号
\usepackage{array}      % 表格扩展支持
\usepackage{booktabs}   % 提供更好的表格线条
\usepackage{multirow}   % 支持单元格多行合并
\usepackage{caption}    % 自定义图片说明格式
\usepackage[utf8]{inputenc}
\usepackage[left=3cm, right=3cm]{geometry}  % 页面边距设置
\usepackage{csquotes}   % 引用格式
\usepackage[normalem]{ulem}  % 删除线等效果
\usepackage{tabularx}   % 表格扩展
\usepackage{wrapfig}    % 图片文字环绕

\usepackage{tikz}       % 支持绘图
\usetikzlibrary{shapes.geometric, arrows, calc, positioning} % tikz扩展库

\usepackage[colorlinks=true, linkcolor=black, citecolor=green, urlcolor=blue]{hyperref}

\usepackage[style=numeric, sorting=none]{biblatex}
\DeclareFieldFormat{eprint}{}
\DeclareFieldFormat{eprinttype}{}
\DeclareFieldFormat{eprintclass}{}

\usepackage{indentfirst}

\renewcommand{\arraystretch}{1.2}  % 调整行高
\usepackage[justification=centering]{caption}  % 确保图注文字居中

\usepackage[english]{babel}

\usepackage{float}  % 控制图片位置

\tikzstyle{block} = [rectangle, draw, fill=blue!20, text centered, minimum height=2em, minimum width=3cm]
\tikzstyle{blockgreen} = [rectangle, draw, fill=green!20, text centered, minimum height=2em, minimum width=3cm]
\tikzstyle{circleblock} = [circle, draw, text centered, minimum size=1.5em]
\tikzstyle{line} = [draw, -latex']
\tikzstyle{group} = [dashed, thick, draw=blue, rounded corners]% 

\usepackage{authblk}  % 用于作者和机构管理
\usepackage{hyperref} % 用于处理超链接

\title{Binary Neural Networks for Large Language Model: A Survey}
\author[1,2]{Liangdong Liu}
\author[1]{Zhitong Zheng\thanks{Corresponding author: \texttt{liam@oppo.com}}}
\author[]{Cong Wang}
\author[]{Tianhuang Su}
\author[]{Zhenyu Yang}
\affil[1] {OPPO AI Center}
\affil[2] {School of
Electrical and Automation Engineering, Nanjing Normal University}
\affil[ ]{\texttt{yh678688@163.com, \{liam, wangcong12,sutianhuang, yangzhenyu\}@oppo.com}}

\date{}  % 去掉日期，或根据需要设置日期
\begin{document}

\maketitle
\begin{center}
\textbf{\Large Abstract}
\end{center}

Large language models (LLMs) have wide applications in the field of natural language processing(NLP), such as GPT-4 and Llama. However, with the exponential growth of model parameter sizes, LLMs bring significant resource overheads. Low-bit quantization, as a key technique, reduces memory usage and computational demands by decreasing the bit-width of model parameters, activations, and gradients. Previous quantization methods for LLMs have largely employed Post-Training Quantization (PTQ) and Quantization-Aware Training (QAT). PTQ does not require any retraining of the original model, while QAT involves optimizing precision during training to achieve the best quantization parameters. The BitNet team proposed a radically different approach, where quantization is performed from the start of model training, utilizing low-precision binary weights during the training process. This approach has led to the emergence of many binary quantization techniques for large language models. This paper provides a comprehensive review of these binary quantization techniques. Specifically, we will introduce binary quantization techniques in deep neural networks and further explore their application to LLMs, reviewing their various contributions, implementations, and applications.


\section{Introduction}

With the rapid development of large language models (LLMs) in the field of natural language processing (NLP) \cite{brown2020language,openai_gpt4,chowdhery_palm,anil_palm2,touvron_llama_v1,touvron_llama2_v2,openai_gpt4_v2,touvron_llama_v3,dubey_llama3_v4,lozhkov_starcoder2,liu_deepseek_v2}, these models have demonstrated exceptional performance in language generation, text understanding, and task reasoning. However, along with the exponential growth in model parameter sizes, LLMs also introduce significant resource overheads, including high memory usage, computational complexity, and increased energy consumption, which present substantial challenges for the deployment and practical application of these models. To address these challenges, low-bit quantization has emerged as a crucial technique for enhancing the efficiency and deployability of LLMs. Most previous research in this field has primarily focused on Post-Training Quantization (PTQ) \cite{yao_zeroquant,frantar_gptq,xiao_smoothquant} and Quantization-Aware Training (QAT) \cite{liu_llm_qat,chen2024efficientqat}. PTQ allows a trained FP32 model to be directly converted into a fixed-point computation model without requiring retraining of the original model. QAT, on the other hand, involves quantizing a trained model and then retraining it.

In this paper, we primarily focus on binary quantization techniques, with particular emphasis on the BitNet \cite{wang_bitnet} method, which first implemented a binary quantization approach for large language models. This method is distinct from PTQ and QAT because BitNet performs quantization from the outset of model training, achieving high energy efficiency in training and inference through binary weights. Since the introduction of this novel approach, many studies \cite{ma_era_1bit,wang_bitnet_a48,brickner_noise_step,ma_fbi_llm,tang_bi_mamba} have explored ways to improve the accuracy of binary large language models and how to implement them on low-power, resource-constrained platforms. This paper provides a comprehensive review of these studies.

In this paper, we will provide a detailed explanation of the concept of binarization and its applications across various domains. We will also thoroughly explain the development of binary quantization techniques from traditional deep neural networks to the field of LLMs, with a particular focus on the research on binary quantization in large language models.

\section{Background}

Previous quantization methods for large language models (LLMs) include PTQ \cite{yao_zeroquant,frantar_gptq,xiao_smoothquant} and QAT \cite{liu_llm_qat,chen2024efficientqat}. However, both methods suffer from significant precision loss at lower bit widths. Yi Guo et al. \cite{guo2024decoupleq} demonstrated the results of various PTQ methods on the Llama-1/2 models using the WikiText-2 dataset, as shown in Table \ref{tab:table 1}.

\begin{table}[ht]
\centering
\small % 设置字体较小
\renewcommand{\arraystretch}{0.8}
\begin{tabular}{@{}lccccccccc@{}}
\toprule
\textbf{Llama}  &  & \textbf{1-7B} & \textbf{1-13B} & \textbf{1-30B} & \textbf{1-65B} & \textbf{2-7B} & \textbf{2-13B} & \textbf{2-70B} \\ \midrule
FP16          &    & 5.68          & 5.09           & 4.10           & 3.53           & 5.47          & 4.88           & 3.31           \\ \midrule
\multirow{4}{*}{W2A16} 
                  & GPTQ          & 2.1e3          & 5.5e3          & 499.75         & 55.91         & 7.7e3          & 2.1e3          & 77.95          \\
                  & OmniQuant     & 15.47          & 13.21          & 8.71           & 7.58          & 37.37          & 17.21          & 7.81           \\
                  & \textbf{decoupleQ} & \textbf{9.49} & \textbf{7.86} & \textbf{6.37}  & \textbf{5.59}  & \textbf{9.74}  & \textbf{13.03} & \textbf{5.23}  \\
                  & runtime       & 2.5            & 4.8            & 12.7           & 27.6          & 2.5            & 4.5            & 33.4           \\ \midrule
\multirow{4}{*}{W2A16g128} 
                  & GPTQ          & 44.01          & 15.60          & 10.92          & 9.51          & 36.77          & 28.14          & -              \\
                  & OmniQuant     & 8.90           & 7.34           & 6.59           & 5.65          & 9.62           & 7.56           & 6.11           \\
                  & \textbf{decoupleQ} & \textbf{8.65} & \textbf{7.25} & \textbf{6.04}  & \textbf{5.19}  & \textbf{8.79}  & \textbf{7.44}  & \textbf{4.96}  \\
                  & runtime       & 3.7            & 7.7            & 24.3           & 55.0          & 3.7            & 7.9            & 70.6           \\ \midrule
\multirow{4}{*}{W2A16g64} 
                  & GPTQ          & 22.10          & 10.06          & 8.54           & 8.31          & 20.85          & 22.44          & -              \\
                  & OmniQuant     & 8.90           & 7.34           & 6.59           & 5.65          & 9.62           & 7.56           & 6.11           \\
                  & \textbf{decoupleQ} & \textbf{8.18} & \textbf{6.96} & \textbf{5.81}  & \textbf{5.07}  & \textbf{8.41}  & \textbf{6.98}  & \textbf{5.34}  \\
                  & runtime       & 4.3            & 8.9            & 27.9           & 64.5          & 4.4            & 9.0            & 98.2           \\ \midrule
\multirow{4}{*}{W3A16} 
                  & GPTQ          & 8.06           & 6.76           & 5.84           & 5.06          & 8.37           & 6.44           & 4.82           \\
                  & AWQ           & 11.88          & 7.45           & 10.07          & 5.21          & 24.00          & 10.45          & -              \\
                  & OmniQuant     & 6.49           & 5.68           & 4.74           & \textbf{4.04}          & 6.58           & \textbf{5.58}           & 3.92           \\
                  & \textbf{decoupleQ} & \textbf{6.38} & \textbf{5.60} & \textbf{4.67}  & 6.05  & \textbf{6.22}  & 5.72  & \textbf{3.84}  \\
                  \midrule
\multirow{4}{*}{W4A16} 
                  & GPTQ          & 6.13           & 5.40           & 4.48           & 3.83          & 5.83           & 5.13           & 3.58           \\
                  & AWQ           & 6.08           & 5.34           & 4.39           & 3.76          & 6.15           & 5.12           & -              \\
                  & OmniQuant     & 5.86           & 5.21           & 4.25           & 3.71          & 5.74           & \textbf{5.02}           & 3.47           \\
                  & \textbf{decoupleQ} & \textbf{5.85} & \textbf{5.21} & \textbf{4.24}  & \textbf{3.67}  & \textbf{5.70}  & 5.06  & \textbf{3.45}  \\
\bottomrule
\end{tabular}
\caption{The PPL results of Llama-1/2 on wikitext-2. The runtime represents the time (in hours) for W2 quantization via decoupleQ, and this table is obtained from \cite{guo2024decoupleq}.}
\label{tab:table 1}
\end{table}

From this table, it is evident that the precision loss is quite severe in W2A16 for PTQ. Additionally, in the EfficientQAT proposed by Mengzhao Chen et al. \cite{chen2024efficientqat}, a table was presented comparing the performance of various quantization methods on Llama 2 and 3, evaluating their average zero-shot accuracy across five common sense reasoning tasks, as shown in Table \ref{tab:Table 2}.
\begin{table}[h!]
    \centering
    \renewcommand{\arraystretch}{0.8}
    \small
    \resizebox{\textwidth}{!}{%
    \begin{tabular}{lcccccccccc}
        \toprule
        \textbf{Method} & \textbf{Bits} & \textbf{Type} & \textbf{Group (code)} & \textbf{2-7} & \textbf{2-13} & \textbf{2-70} & \textbf{3-8} & \textbf{3-70} \\
        \midrule
        FP16 & 16 & - & 16 & 64.86 & 67.81 & 72.41 & 68.58 & 75.33 \\
        \midrule
        RTN & 4 & uniform & 128 & 64.52 & 67.50 & 72.26 & 67.79 & 73.98 \\
        GPTQ & 4 & uniform & 128 & 64.24 & 67.27 & 72.39 & 67.80 & 74.74 \\
        AWQ & 4 & uniform & 128 & \textbf{64.54} & \textbf{67.61} & 72.44 & 68.24 & \textbf{74.77} \\
        OmniQ & 4 & uniform & 128 & 64.52 & 67.10 & 72.39 & - & - \\
        AutoRound & 4 & uniform & 128 & 64.39 & 67.36 & 72.47 & - & - \\
        QuIP\# & 4 & vector & - & 64.48& 67.28 & 72.17 & - & - \\
        EfficientQAT & 4 & uniform & 128 & 64.27 & 67.52 & \textbf{72.62} & \textbf{68.43} &74.57\\
        \midrule
        RTN & 3 & uniform & 128 & 62.06 & 65.77 & 70.83 & 58.72 & 65.29 \\
        GPTQ & 3 & uniform & 128 & 62.48 & 66.18 & 71.47 & 60.58 & 71.28 \\
        AWQ & 3 & uniform & 128 & 62.82 & 66.14 & 71.41 & 64.82 & 73.65 \\
        OmniQ & 3 & uniform & 128 & 62.42 & 66.18 & 71.07 & - & - \\
        AutoRound & 3 & uniform & 128 & 63.72 & 66.68 & 71.24 & - & - \\
        QuIP\# & 3 & vector & - & 63.52 & 66.26 & 72.13 & - & - \\
        EfficientQAT & 3 & uniform & 128 & \textbf{64.02} & \textbf{67.28} & \textbf{71.76} & \textbf{67.35} & \textbf{72.42} \\
        \midrule
        OmniQ & 2 & uniform & 128 & 46.98 & 53.56 & 54.87 & - & - \\
        AutoRound & 2 & uniform & 128 & 54.50 & 60.72 & 67.70 & - & - \\
        EfficientQAT & 2 & uniform & 128 & \textbf{59.50} & \textbf{63.88} & \textbf{68.93} & \textbf{59.37} & \textbf{67.57} \\
        AQLM & 2 & vector & 2x8 & 57.61 & 62.22 & 69.85 & - & - \\
        AQLM & 2 & vector & 1x16 & 61.85 & 64.95 & 70.84 & 64.10 & 70.10 \\
        QuIP\# & 2 & vector & - & 60.61 & 64.44 & 70.91 & - & - \\
        EfficientQAT & 2 & uniform & 64 & \textbf{60.14} & \textbf{64.48} & \textbf{69.48} & \textbf{60.76} & \textbf{67.89} \\
        \bottomrule
    \end{tabular}%
    }
    \caption{The average zero-shot accuracy ($\uparrow$) of Llama 2 and 3 on 5 common-sense reasoning tasks. "Group" indicates the group size for uniform quantization and the codebook scheme for vector quantization. "Bold" represents the best results for uniform quantization. This table is sourced from \cite{chen2024efficientqat}.}
    \label{tab:Table 2}
\end{table}
Similarly, both PTQ and QAT perform poorly when the bit width is reduced to 2. The precision loss for PTQ is notably severe, while QAT maintains approximately 7\% accuracy loss at 2 bits, which is unacceptable in industrial applications. Therefore, it is clear that traditional quantization methods are not feasible when further quantizing weights, such as in the case of 1-bit quantization. However, in the CNN field, 1-bit quantization trained from scratch has matured and achieved satisfying results, providing a valuable reference for researchers seeking to implement 1-bit quantization in LLMs. Yiwei Lu et al. \cite{lu2024understanding} provided Table \ref{tab:Table 3}, which presents the accuracy of various BNN-CNN models on CIFAR-10 and ImageNet-1k. From Table \ref{tab:Table 3}, it is evident that this approach to 1-bit quantization from scratch can maintain high accuracy in CNNs, while significantly reducing storage costs due to the 1-bit quantization.

\begin{table}[ht]
\centering
\small % 缩小字体以适应表格
\setlength{\tabcolsep}{3pt} % 调整列间距
\renewcommand{\arraystretch}{0.8} % 轻微减少行间距
\resizebox{\linewidth}{!}{%
\begin{tabular}{@{}l l l c c c c c c c c@{}}
\toprule
\multirow{2}{*}{\textbf{Dataset}} & \multirow{2}{*}{\textbf{Pipeline}} & \multirow{2}{*}{\textbf{Task}} & \multirow{2}{*}{\textbf{FP}} & \multirow{2}{*}{\textbf{PQ}} & \multirow{2}{*}{\textbf{rPC}} & \multicolumn{5}{c}{\textbf{ProxConnect++}} \\ 
\cmidrule(lr){7-11}
& & & & & & \textbf{BC} & \textbf{PC} & \textbf{BNN} & \textbf{BNN+} & \textbf{BNN++} \\ 
\midrule
\multirow{6}{*}{CIFAR-10} 
& \multirow{3}{*}{FT} 
& BW  & 92.01\% & 89.94\% & 89.98\% & 90.31\% & 90.31\% & 90.35\% & 90.27\% & \textbf{90.40\%} \\
& & BWA & 92.01\% & 88.79\% & 83.55\% & 89.39\% & 89.95\% & 90.01\% & 89.99\% & \textbf{90.22\%} \\
& & BWAA & 92.01\% & 85.39\% & 81.10\% & 89.11\% & 89.21\% & 89.32\% & 89.55\% & \textbf{90.01\%} \\ 
\cmidrule(lr){2-11}
& \multirow{2}{*}{E2E} 
& BW  & 92.01\% & 81.59\% & 81.82\% & 87.51\% & 88.05\% & 89.92\% & 89.39\% & \textbf{90.03\%} \\
& & BWA & 92.01\% & 81.51\% & 81.60\% & 86.99\% & 87.26\% & 89.15\% & 89.02\% & \textbf{89.91\%} \\ 
\midrule
\multirow{6}{*}{ImageNet-1K} 
& \multirow{3}{*}{FT} 
& BW  & 78.87\% & 66.77\% & 69.22\% & 71.35\% & 71.29\% & 71.41\% & 70.22\% & \textbf{72.33\%} \\
& & BWA & 78.87\% & 56.21\% & 58.19\% & 65.99\% & 65.61\% & 66.02\% & 65.22\% & \textbf{68.03\%} \\
& & BWAA & 78.87\% & 53.29\% & 55.28\% & 58.18\% & 59.21\% & 59.77\% & 59.10\% & \textbf{63.02\%} \\ 
\cmidrule(lr){2-11}
& \multirow{2}{*}{E2E} 
& BW  & 78.87\% & 63.23\% & 66.39\% & 67.45\% & 67.51\% & 67.49\% & 66.99\% & \textbf{68.11\%} \\
& & BWA & 78.87\% & 61.19\% & 64.17\% & 65.42\% & 65.31\% & 65.29\% & 65.98\% & \textbf{66.08\%} \\ 
\bottomrule
\end{tabular}%
}
\caption{
The performance of different ProxConnect++ variants (BC, PC, BNN, BNN+, and BNN++) compared with FP, PQ, and rPC in terms of test accuracy. BW stands for binarizing weights, BWA stands for binarizing weights and activations, and BWAA represents binarizing weights, activations, and 8-bit accumulators. FT stands for fine-tuning, and E2E stands for end-to-end. This table is sourced from \cite{lu2024understanding}.}
\label{tab:Table 3}
\end{table}

Inspired by BNN-CNN, researchers began to explore whether this technique could be applied to LLMs as well. This led to the birth of BitNet \cite{wang_bitnet}, the pioneering work in BNN-LLM. The BitNet team, not satisfied with the results from BitNet, introduced BitNet b1.58 \cite{ma_era_1bit}, which showed promising performance. This led to a series of related studies, including BitNet a4.8 \cite{wang_bitnet_a48}, FBI-LLM \cite{ma_fbi_llm}, and Bi-Mamba \cite{tang_bi_mamba}.

Currently, BitNet has been successfully applied in CPU environments achieving high efficiency \cite{wei2024tmac}. Furthermore, BitNet has been extended to the multimodal domain \cite{sundaram2024llavaolmobitnet1b}, showcasing its strong competitiveness. These developments suggest a promising future for 1-bit quantization technology in LLMs, making it possible to develop low-cost yet high-precision LLMs. In the following section, we will introduce several classic BNN-CNN algorithms and further discuss the emergence of BNN-LLM.

\section{Binary Neural Networks for Convolutional Neural Networks}
In the field of deep neural networks (DNNs), many studies have focused on making models smaller and faster without significantly sacrificing accuracy \cite{howard_mobilenets,iandola_squeezenet,hanson_biases_backpropagation,cun_optimal_brain_damage,han_deep_compression}. In theory, smaller data types not only reduce model size but also improve computational speed, as fixed-point operations are significantly more efficient than floating-point operations. Gupta et al. \cite{gupta_deep_learning_precision} pointed out that reducing the precision of data types in DNNs can decrease model size with an acceptable impact on accuracy. Courbariaux et al. \cite{courbariaux_training_low_precision} compared the training accuracy of DNNs using fixed-point and floating-point values of various bit-widths. They also examined the effectiveness of a hybrid dynamic fixed-point data type, demonstrating that comparable accuracy can be achieved with precision below 32 bits.

\begin{figure}[ht]  % 创建浮动图形环境
    \centering
    \includegraphics[page=1, trim=20 160 135 10, clip, width=\textwidth]{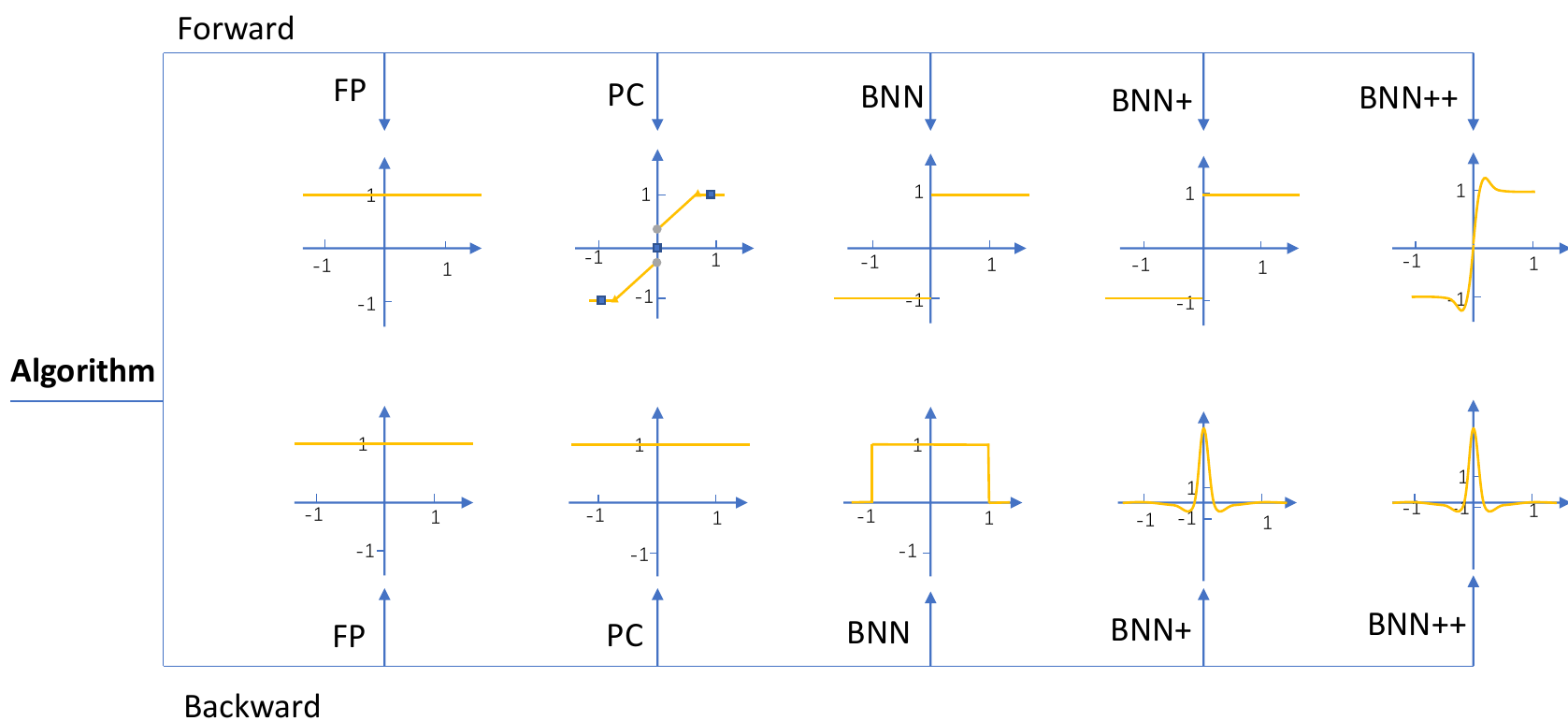}
    \caption{The forward and backward propagation diagrams of different algorithms, where the upper/lower part represents forward/backward propagation.}  % 图说明
    \label{fig:Figure1}  % 图标签，可以用于引用
\end{figure}

Courbariaux et al. \cite{courbariaux2016binarynet,hubara2016binarized} pioneered the BNN methodology, which forms the foundation and starting point for most BNN-CNN algorithms and research in the field. Subsequently, a variety of binary neural network algorithms have been proposed \cite{courbariaux_binaryconnect,bengio2013estimating,rastegari2016xnor,zhou2016dorefa,tang2017compact,lin2017binary,darabi2018bnnplus,dockhorn2021demystifying,hubara2016binarized,lu2024understanding,hubara2016binarized_v1,prabhu2018hybrid,lu2024understanding,tang2017compact,wang2018flexible,zhao2017accelerating,fraser2017scaling,guo2018fbna,hubara_binarized_nns,seo2016binarizing,yonekawa2018ternary,hwang2014fixed,prost2018high}. Among them, BinaryConnect (BC) \cite{courbariaux_binaryconnect} is considered the standard. BC binarizes weights using the sign function during forward propagation and uses the Straight-Through Estimator (STE) to evaluate the gradients of the binarized weights. This idea has also been widely adopted in the LLM field, which we will introduce in Chapter 4. Based on BC, Dockhorn et al. \cite{dockhorn2021demystifying} further proposed ProxConnect (PC), which extends BC by allowing the use of proximal quantizer during forward propagation, with the sign function being a special case. Additionally, Courbariaux et al.'s paper \cite{courbariaux2016binarynet} first introduced the method of binarizing activation values. Darabi et al. \cite{darabi2019regularized} further improved BNN by using the derivative of the Sign-Swish (SS) function as the backward quantizer, while retaining the sign function as the forward quantizer. BNN++ \cite{lu2024understanding} is an extension of BNN+ \cite{darabi2019regularized}, where the sign forward quantizer is replaced with the Sign-Swish (SS) function, further optimizing the binary neural network algorithm. A comparison of the forward and backward quantizers of various classic binary neural network algorithms is shown in Figure \ref{fig:Figure1} and Table \ref{tab:Table 4}.

\begin{table}[ht]
\centering
\renewcommand{\arraystretch}{0.8}
\small
\begin{tabular}{|c|c|c|}
\hline
\textbf{Forward Quantizer} & \textbf{Backward Quantizer} & \textbf{Algorithm} \\
\hline
identity & identity & FP \\
\hline
$P_{Q}$ & identity & BC \\
\hline
$ L^\varrho_\rho
\ $ & identity & PC \\
\hline
$P_{Q}$ & $1_{[-1,1]}$ & BNN \\
\hline
$P_{Q}$ & $\nabla SS$ & BNN+ \\
\hline
SS & $\nabla SS$ & BNN++ \\
\hline
\end{tabular}
\caption{Variants of ProxConnect++. This table is taken from \cite{lu2024understanding}.}
\label{tab:Table 4}
\end{table}

As can be seen, binary quantization technology has already matured in the traditional CNN field. Therefore, in the LLM field, after researchers noticed that traditional PTQ and QAT methods suffer from significant precision loss when the bit width is low, such as at 2 bits, they shifted their focus towards the 1-bit quantization technology trained from scratch. In the following section, we will provide a detailed review of these studies.

\section{Binary Neural Networks for Large Language Models}
Previous quantization techniques for large language models existed \cite{lin_awq,frantar_gptq,shi_inferflow,xia_fp6_llm,dettmers_spqr,zhang_qqq,lee_tender,dong_qaq,liu_intactkv,kang_gear,lin_qserve,lin_qserve,liu_kivi,yang_no_token_left,yao_zeroquant,dettmers_4bit_precision,frantar_gptq,liu_bi_real_net,qin_bibert,liu_llm_qat}, but unlike these techniques, Hongyu Wang et al. \cite{wang_bitnet} drew inspiration from algorithms in BNN-CNN and pioneered the application of binarization to large language models, introducing BitNet, which forms the foundation and starting point for most of the research on binary quantization of large language models. In this chapter, we will focus on reviewing the binarization techniques specifically for large language models.

\subsection{The Pioneering Work in BNN-LLM: BitNet}

BitNet, proposed by Hongyu Wang et al. \cite{wang_bitnet}, is a 1-bit Transformer architecture for large language models that achieves efficient scaling in terms of both memory and computation. BitNet introduces a novel linear projection layer, BitLinear, which replaces the standard Linear layer in the Transformer model to binarize the weights. Specifically, BitLinear substitutes traditional matrix multiplication with a 1-bit weight version.

Following the principles of BNNs, BitLinear first binarizes the weights using the sign function, assigning them values of +1 or -1. In line with the ideas of Zechun Liu et al. \cite{liu_bit_neurips}, Hongyu Wang et al. \cite{wang_bitnet} adjust the mean of the weights to zero before binarization in order to enhance the capacity within the limited numerical range. At this point, the binarization process of a weight matrix \( W \) can be represented as:
\begin{equation}
\widetilde{W} = \text{Sign}(W - \alpha)
\end{equation}
Here:
\begin{equation}
\text{Sign}(W_{ij}) =
\begin{cases} 
+1, & \text{if } W_{ij} > 0, \\
-1, & \text{if } W_{ij} \leq 0,
\end{cases}
\end{equation}
\begin{equation}
\alpha = \frac{1}{nm} \sum_{ij} W_{ij}
\end{equation}

Following the method of Tim Dettmers et al. \cite{dettmers_llm_int8}, Hongyu Wang et al. used absolute maximum (Absmax) quantization to binarize the activation values to b-bit precision. To maintain the variance unchanged after quantization, they introduced the LayerNorm \cite{ba_layer_norm} function before activation quantization. At this point, the estimated variance of the output \( y \) is given by 

\begin{equation}
\text{Var}(y) \approx E\left[\text{LN}(\widetilde{X})^2\right] = 1,
\end{equation}

which ensures that its magnitude is the same as the variance of the full-precision counterpart. This implementation is equivalent to the SubLN \cite{wang_foundation_transformers} in transformers. The activation values are scaled by multiplying by \( Q_b \) and dividing by the maximum value of the input matrix, ensuring that the activation values fall within the range \([-Q_b, Q_b]\). The formula is as follows:

\begin{equation}
\widetilde{x} = \text{Quant}(x) = \text{Clip}\left(x \times \frac{Q_b}{\gamma}, -Q_b + \epsilon, Q_b - \epsilon \right)
\end{equation}

\begin{equation}
\text{Clip}(x, a, b) = \max\left(a, \min(b, x)\right), \quad \gamma = \|x\|_\infty
\end{equation}

Here, \( \epsilon \) is a small floating-point number, which is used to prevent overflow during the clipping operation. For activation values before nonlinear functions (e.g., ReLU), they are scaled to the range \([0, Q_b]\). The formula for BitLinear defined at this point is as follows:

\begin{equation}
y = \widetilde{W} \widetilde{x}=\widetilde{W} \text{Quant}(\text{LN}(x)) \times \frac{\beta \gamma}{Q_b}
\end{equation}

\begin{equation}
\text{LN}(x) = \frac{x - E(x)}{\sqrt{\text{Var}(x) + \epsilon}}, \quad \beta = \frac{1}{nm} \|W\|_1
\end{equation}

BitNet performs 8-bit quantization on the activation values, with quantization applied to each tensor during training and to each token during inference, ensuring both stability and efficiency.

Hongyu Wang et al. \cite{wang_bitnet} proposed dividing the weights and activations into multiple groups and independently computing the parameters for each group. This approach allows for local computation of parameters without the need for additional communication operations. The formula is as follows:

\begin{equation}
\alpha_g = \frac{G}{nm} \sum_{ij} W_{ij}^{(g)}, \quad \beta_g = \frac{G}{nm} \|W^{(g)}\|_1
\end{equation}

Here, \( W^{(g)} \) represents the weights of the g-th group. Similarly, for the activation values, we can divide the input matrix \( x \in R^{n \times m} \) into \( G \) groups and compute the parameters for each group as follows:

\begin{equation}
\gamma_g = \|x^{(g)}\|_\infty, \quad \eta_g = \min_{ij} x_{ij}^{(g)}
\end{equation}

\begin{figure}[ht]  % 创建浮动图形环境
    \centering
    \includegraphics[page=1, trim=240 90 240 115, clip, width=\textwidth]{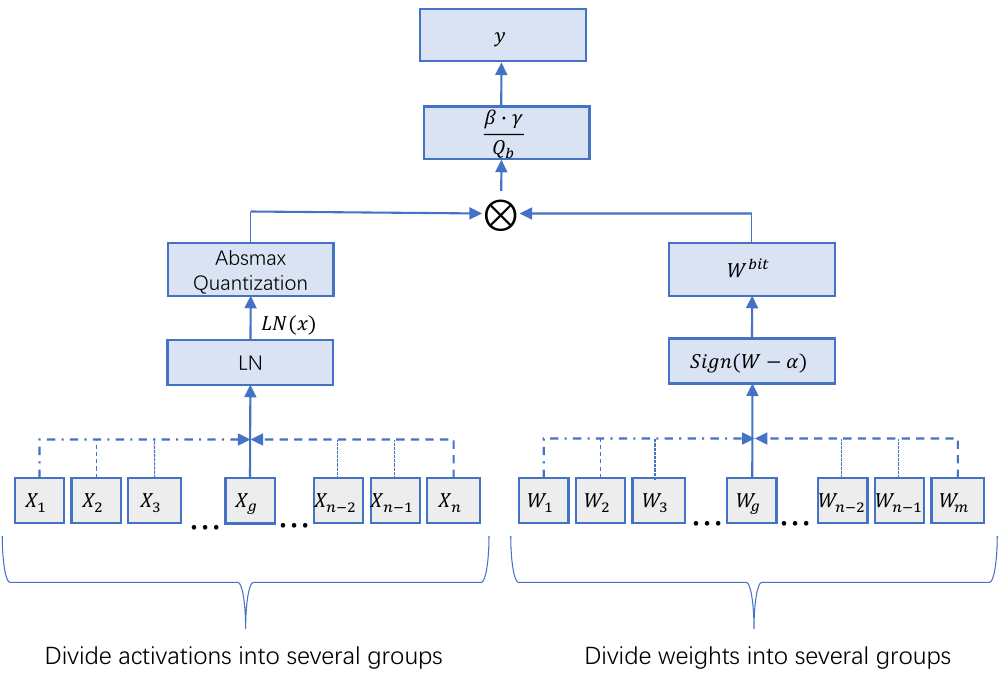}
    \caption{Schematic of BitLinear, where in this diagram, it is assumed that the activations are divided into n groups and the weights are divided into m groups.} % 图注
    \label{fig:Figure 2} % 可选标签，用于引用
\end{figure}

In summary, the schematic diagram of BitLinear is shown in Figure \ref{fig:Figure 2}.

\subsection{Optimized Binary Neural Networks for Large Language Models}

\subsubsection{Optimization of Weight Quantization Methods}

Similar to the weight quantization idea in BitNet, FBI-LLM \cite{ma_fbi_llm} and Bi-Mamba \cite{tang_bi_mamba} also attempt to use the Sign function to binarize the weights to +1 or -1. However, the BitNet network proposed by Hongyu Wang et al. \cite{wang_bitnet} adjusts the binarized weights with a scaling factor $\beta$, whereas FBI-LLM and Bi-Mamba introduce column-based learnable scaling factors $\alpha_j$ and $\beta_j$ to reduce the error between the binarized weights and the full-precision ones. BitNet, FBI-LLM, and Bi-Mamba all attempt to binarize the weights, while BitNet b1.58 \cite{ma_era_1bit} uses absolute mean (absmean) quantization to ternarize the weights to $\{-1, 0, 1\}$. BitNet b1.58 points out that its modeling capability is enhanced by its explicit support for feature filtering, which is enabled by incorporating 0 values in the model weights, thereby significantly boosting the performance of 1-bit LLMs.

\begin{figure}[ht] 
    \centering
    \includegraphics[page=1, trim=216 160 190 80, clip, width=\textwidth]{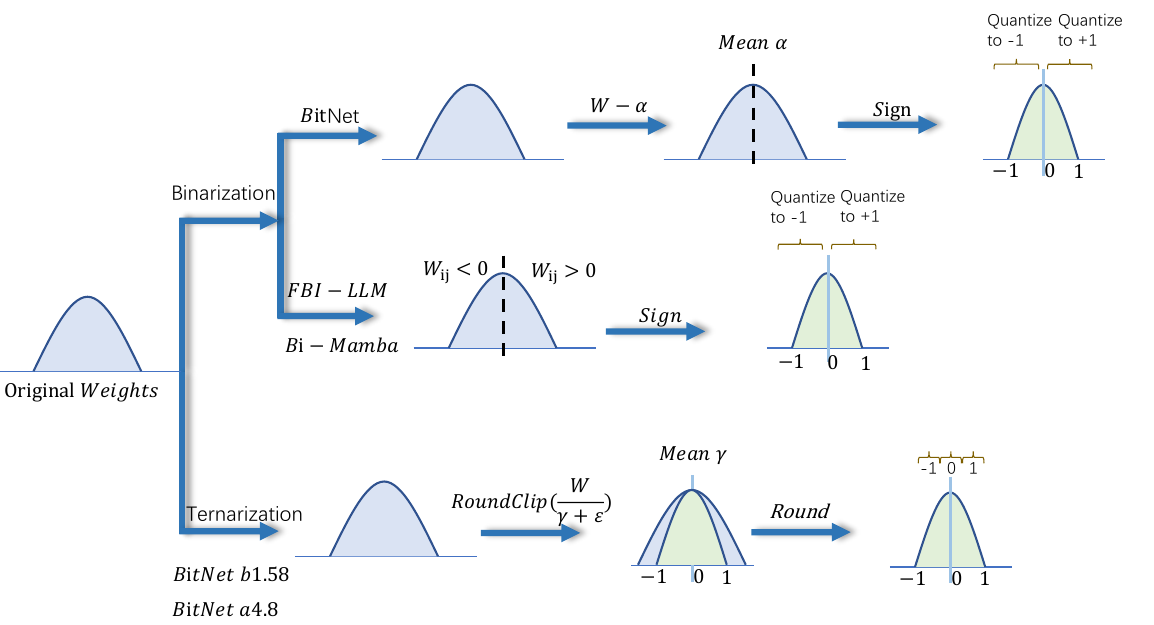}
    \caption{A comparison of weight quantization for different networks, where the blue/green plots represent the results before/after quantization.} 
    \label{fig:Figure 3}  % 规范标签命名
\end{figure}

\begin{figure}[ht] 
    \centering
    \includegraphics[page=1, trim=60 30 25 10, clip, width=\textwidth]{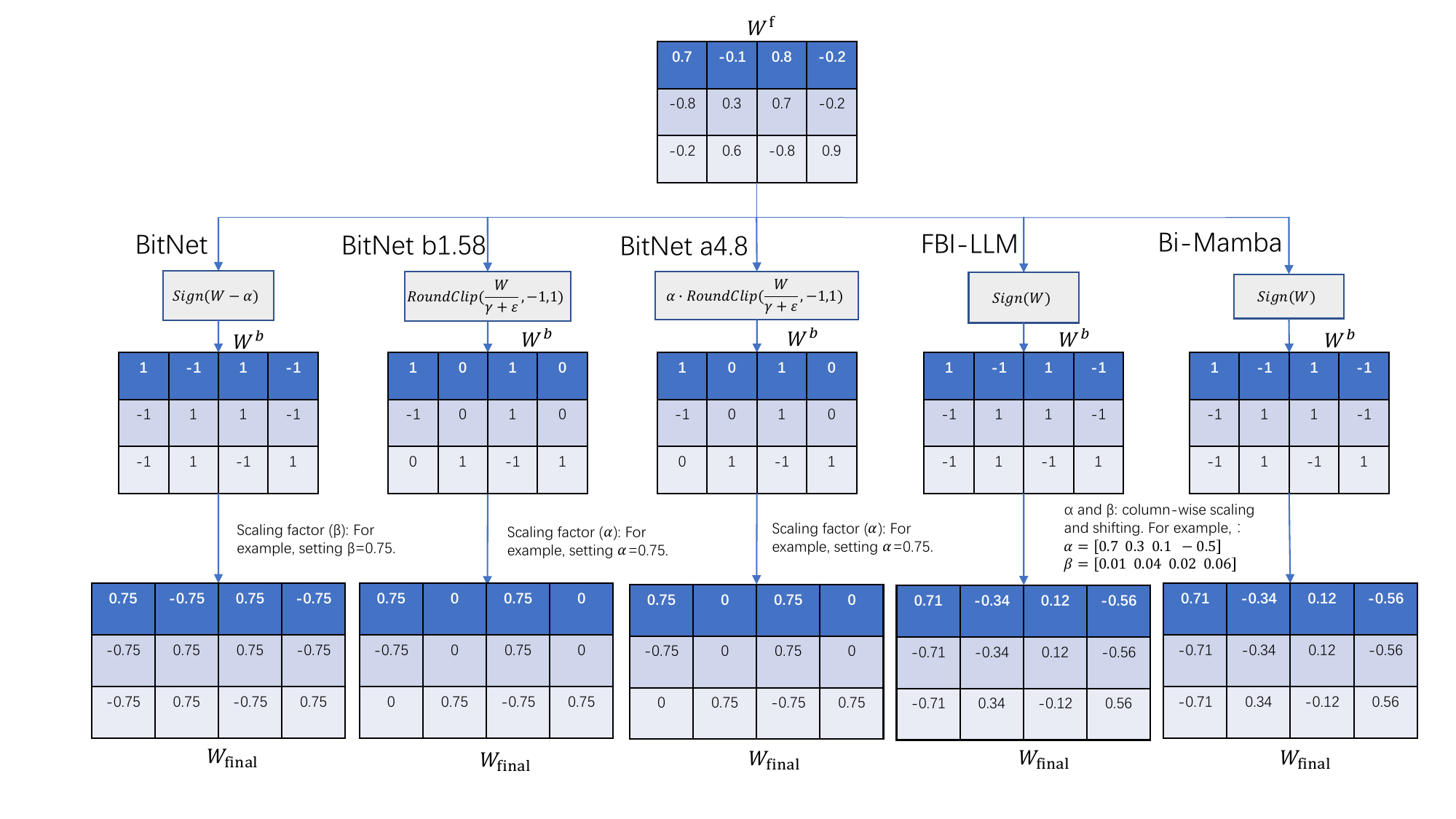 } % 替换为你的 PDF 文件名
    \caption{A specific example of weight quantization methods across different networks. The scaling factors for BitNet, BitNet b1.58, and BitNet a4.8 are set to 0.75, while the scaling factors for FBI-LLM and Bi-Mamba are set as follows: \( \alpha = [0.7, 0.3, 0.1, -0.5] \) and \( \beta = [0.01, 0.04, 0.02, 0.06] \).
} 
    \label{fig:Figure 4}  % 规范标签命名
\end{figure}

BitNet a4.8 \cite{wang_bitnet_a48} adopts the same architecture as BitNet b1.58, learning 1.58-bit weights from scratch, but with further compression applied to the activations, which we will discuss in subsequent chapters. Based on the above description, the comparison of weight quantization across different networks is shown in Figure \ref{fig:Figure 3}, while a specific example related to this comparison is presented in Figure \ref{fig:Figure 4}.

\subsubsection{Optimization of Activation Quantization Methods}

In Section 4.1, we provided a detailed explanation of the activation method in BitNet \cite{wang_bitnet}. While BitNet b1.58 \cite{ma_era_1bit} quantizes the weights to three values, it also makes certain changes to the activations. Specifically, BitNet scales the activation values to the range \([0, Q_b]\) before the nonlinear function, while BitNet b1.58 scales the activations to \([-Q_b, Q_b]\) for each token. This design helps avoid zero-point quantization. The activation values in both BitNet and BitNet b1.58 are 8 bits. 

In contrast, BitNet a4.8 \cite{wang_bitnet_a48} introduces several changes to activation quantization. The BitNet a4.8 team incorporates hybrid quantization and sparsification strategies to support 4-bit activations in 1-bit LLMs, addressing quantization errors. They use 4-bit activations for the inputs to the attention and feed-forward networks while sparsifying intermediate states and quantizing them to 8 bits. BitNet a4.8 is trained from 8-bit activations to 4-bit activations using a two-stage training strategy. Compared to BitNet b1.58, BitNet a4.8 achieves faster inference speeds.

\begin{figure}[ht]
    \centering
    \includegraphics[page=1, trim=0 0 33 60, clip, width=\textwidth]{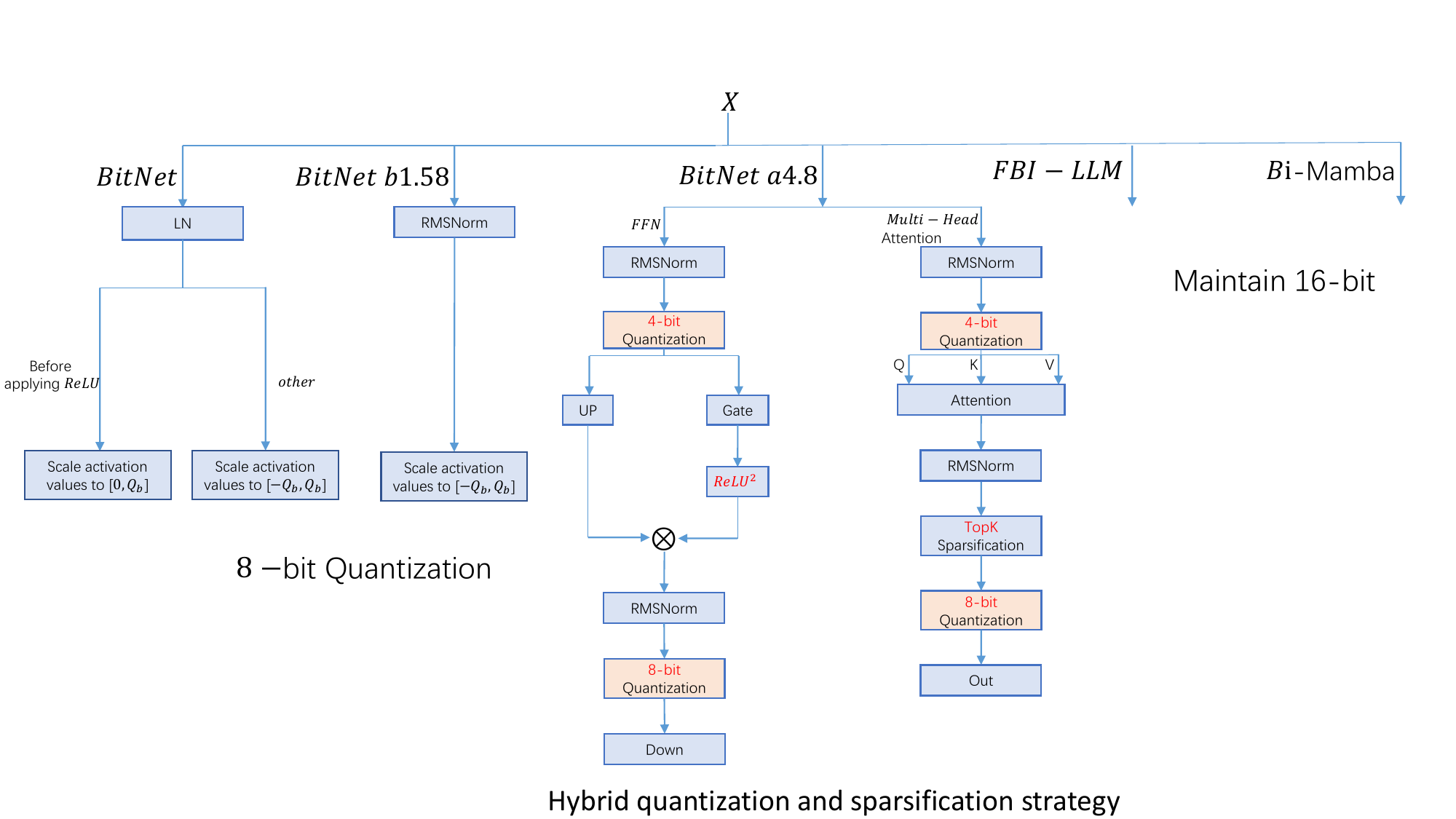} % 替换为你的 PDF 文件名
    \caption{A comparison of activation quantization techniques across various networks. BitNet a4.8 utilizes a hybrid quantization and
sparsification strategy to address outlier activations in certain Transformer sub-layers.} % 图注
    \label{fig:Figure 5} % 可选标签，用于引用
    
\end{figure}

Although BitNet b1.58, proposed by the BitNet team, significantly improves the performance of LLMs, we observe that its training is 60\% slower than the baseline and relatively more costly. This is because RMSNorm has been activated. Considering this, FBI-LLM \cite{ma_fbi_llm} and Bi-Mamba \cite{tang_bi_mamba} only quantize the weights, leaving the activations as 16-bit. Figure \ref{fig:Figure 5} shows a comparison of activation quantization in BitNet \cite{wang_bitnet}, BitNet b1.58 \cite{ma_era_1bit}, BitNet a4.8 \cite{wang_bitnet_a48}, FBI-LLM \cite{ma_fbi_llm}, and Bi-Mamba \cite{tang_bi_mamba}.

\subsubsection{KV Cache Quantization}
\begin{figure}[ht]
    \centering
    \includegraphics[page=1, trim=190 70 195 90, clip, width=\textwidth]{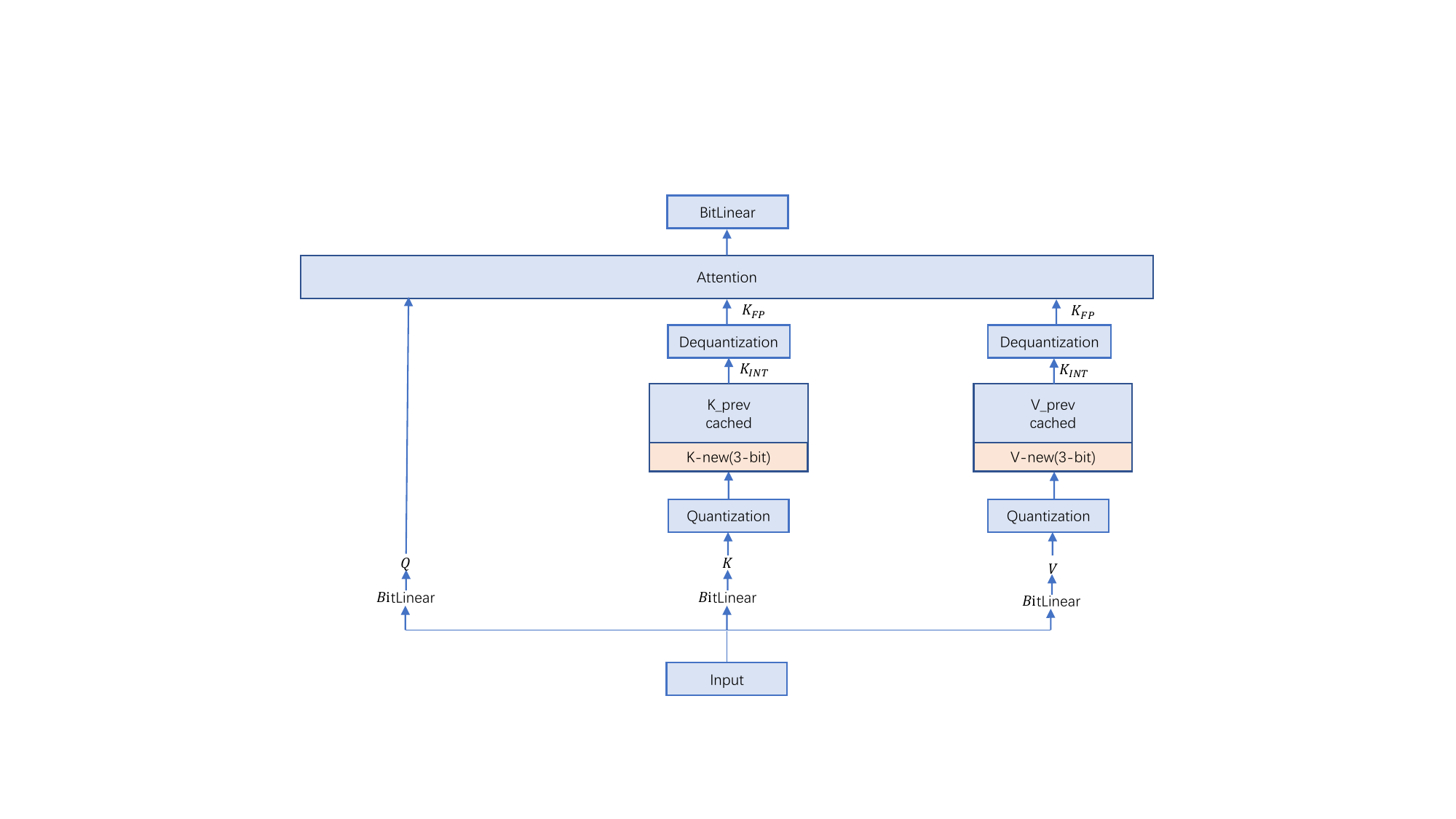} % 替换为你的 PDF 文件名
    \caption{The schematic of KV cache quantization in BitNet a4.8, where K and V are quantized first and then dequantized.} % 图注
    \label{fig:Figure 6} % 可选标签，用于引用
    
\end{figure}

During large model inference, the model's memory usage is primarily determined by the model weights, activations, and KV cache. KV cache quantization is mainly divided into two categories: full quantization \cite{sheng_flexgen,yue_wkvquant} and KV cache-only quantization \cite{hooper_kvquant,dong_get_more_with_less,yang_no_token_left,dong_qaq,kang_gear}. BitNet and BitNet b1.58 focus solely on quantizing the weights and activations, while FBI-LLM and Bi-Mamba focus only on weight quantization, without considering KV cache quantization. However, BitNet a4.8 \cite{wang_bitnet_a48}, proposed by the BitNet team, explicitly states that it supports 3-bit KV cache. Specifically, BitNet a4.8 uses post-RoPE quantization. Their QKV heads are directly quantized to unsigned integers using the absmax function, without the need for a calibration dataset. The schematic diagram of KV cache quantization in BitNet a4.8 is shown in Figure  \ref{fig:Figure 6} .

\subsubsection{Improved Network Loss Function}
Minimizing quantization errors typically attempts to retain the values of full-precision weights and activations, thereby reducing the information loss at each layer. However, focusing solely on the precise approximation of local layers makes it difficult to guarantee the accuracy of the output after passing through a series of layers. As a result, many researchers have dedicated efforts to finding the ideal network loss function  \cite{xu_onebit,bhardwaj_oh_we_freeze,pham_collaborative_distillation,boo_stochastic_precision,kim_qkd}.

FBI-LLM \cite{ma_fbi_llm} differs from traditional autoregressive language models by training FBI-LLM through autoregressive distillation (AD). During training, they use a full-precision pre-trained LLM as the teacher model and the binarized target model as the student model. The cross-entropy loss between the output of the student model and the output of the teacher model is computed at each step of predicting the next token and is used as the final loss function. The formula is as follows:

\begin{equation}
L = -\frac{1}{n} \sum_{i}^{n} p^T(x^{i+1}) \cdot \log p^S(x^{i+1})
\end{equation}

Here, $n$ represents the number of input tokens, $p_j^T(x^{i+1})$ denotes the teacher model's predicted distribution over the vocabulary at step $i$, while $p^S(x^{i+1})$ represents the corresponding predicted distribution of the student model. Experiments with FBI-LLM show that for binarized LLMs, using only the distillation loss yields better results than using the traditional one-hot label autoregressive loss, while maintaining the simplicity of the method. However, we found that for FBI-LLM, if the distillation is removed, the performance significantly degrades and may even fail to converge. This observation warrants further research.

\subsection{Training Binary Neural Networks for Large Language Models}

\subsubsection{Optimization of Gradient Calculation Methods}

BitNet \cite{wang_bitnet}, BitNet b1.58 \cite{ma_era_1bit}, BitNet a4.8 \cite{wang_bitnet_a48}, FBI-LLM \cite{ma_fbi_llm}, and Bi-Mamba \cite{tang_bi_mamba} all attempt to use the Straight-Through Estimator (STE) to address gradient issues. However, inspired by \cite{baydin_gradients_without_backpropagation}, Will Brickner et al. \cite{brickner_noise_step} proposed a new algorithm called noise\_step, which enables the model to be trained directly at ternary precision without the need for backpropagation or momentum mechanisms, while still achieving performance comparable to existing methods, such as the Adam optimizer. Their method highlights that traditional training depends on gradient memory for parameter updates, whereas noise\_step estimates gradients using pseudo-random disturbances and the Jacobian Vector Product (JVP), without the need to store or transmit gradient information. For distributed training, the synchronization of gradients and optimizers is typically a performance bottleneck. noise\_step greatly reduces communication costs by encoding each disturbance using ternary symbols (only 1.58 bits).

\subsubsection{Mixed Precision Training}

BitNet \cite{wang_bitnet} utilizes low-precision weights and activations, while gradients and optimizers are still stored in high precision. This approach ensures the stability and accuracy of the training process. FBI-LLM \cite{ma_fbi_llm} replaces all linear modules except for the causal head with FBI-linear, because the causal head directly affects the distribution of the output token at each step. If the causal head were binarized, it would significantly impact the accuracy of the model’s output. Thus, it is necessary to maintain its precision. Additionally, for two core modules in LLMs—embedding and layer norm—the parameters must be kept in full precision. The embedding module contains semantic information for all tokens, and since it is the first layer of the model, it determines the initial representation of the text, so it cannot be binarized. Layer norm directly scales the activation values, and binarizing its parameters would significantly reduce the semantic expressiveness of each layer’s activations. This approach aligns with the philosophy of the BitNet team.

\subsubsection{Learning Rate Selection}

For both BNN-CNN and BNN-LLM, the choice of learning rate is different. For example, Hongyu Wang et al. \cite{wang_bitnet} discovered that increasing the learning rate can accelerate optimization, and BitNet performs well under a high learning rate. Bi-Mamba \cite{tang_bi_mamba}, in binarized models, accelerates the training process by gradually increasing the learning rate. However, it is worth noting that Jacob Nielsen et al. \cite{nielsen_bitnet_reloaded} argue that for small classification models, whether using 1.58-bit or 16-bit weights, larger learning rates do not perform well, even when training from scratch. On the contrary, a smaller learning rate results in better performance. Tang et al. \cite{tang2017compact} suggested that using a smaller learning rate improves the training results of BNNs. The key reason appears to be that large language models have a very large number of parameters, and after binarization, the granularity of each weight update becomes coarser. If a smaller learning rate is set, training will become slower. However, for small language models or CNNs, their parameter scale is smaller, so only smaller learning rates can be used to avoid gradient explosion.

\subsection{Differences Between BNN-LLM and BNN-CNN}

Binarization techniques have yielded satisfactory results in both CNNs and LLMs, but there are certain differences. First, binarized training in CNNs requires multiple epochs to achieve stable convergence, while binarization in LLMs typically requires only one epoch. Secondly, as mentioned in Section 4.3.3, their learning rate settings are different. Finally, there are differences in computational efficiency and deployment. BNN-CNN typically leverages XNOR operations and popcount instructions, significantly accelerating dot product computations, making it suitable for edge devices such as FPGAs and ASICs. In contrast, BNN-LLM uses mixed precision training, so even though the weights are binarized, the Transformer mechanism still requires processing higher-precision inputs and outputs. This makes it more suited for distributed computing environments, such as GPU clusters and TPU architectures.

\section{Evaluation and Discussion}

\subsection{Comparison of BitNet with PTQ Accuracy}

Hongyu Wang et al. \cite{wang_bitnet} compared BitNet with commonly used post-training quantization methods, such as Absmax \cite{dettmers_llm_int8}, SmoothQuant \cite{xiao_smoothquant}, GPTQ \cite{frantar_gptq}, and QuIP \cite{chee_quip}. These post-training quantization methods were applied to the same FP16 Transformer model, with training settings and data kept consistent with BitNet. The BitNet team provided the table, showcasing the comparison between BitNet and PTQ, as shown in Table \ref{tab:Table 5}.

\begin{table}[ht]
\centering
\renewcommand{\arraystretch}{0.8}
\small
\begin{tabular}{@{}>{\centering\arraybackslash}p{1cm}>{\centering\arraybackslash}p{2cm}>{\centering\arraybackslash}p{1cm}>{\centering\arraybackslash}p{1cm}>{\centering\arraybackslash}p{1cm}>{\centering\arraybackslash}p{1cm}>{\centering\arraybackslash}p{1cm}>{\centering\arraybackslash}p{1cm}>{\centering\arraybackslash}p{1cm}@{}}
\toprule
\textbf{WBits} & \textbf{Methods} & \textbf{PTQ} & \textbf{PPL$\downarrow$} & \textbf{WG$\uparrow$} & \textbf{WGe$\uparrow$} & \textbf{HS$\uparrow$} & \textbf{SC$\uparrow$} & \textbf{Avg$\uparrow$} \\ \midrule
- & Random & $\times$ & - & 50.0 & 50.0 & 25.0 & 50.0 & 43.8 \\
16 & Transformer & $\times$ & 15.19 & 66.7 & 54.3 & 42.9 & 67.4 & 57.8 \\
\midrule
\multirow{2}{*}{8} & Absmax &  $\checkmark$ & 21.43 & 60.4 & 52.0 & 38.3 & 62.7 & 53.4 \\
& SmoothQuant & $\checkmark$ & 15.67 & 65.3 & 53.1 & 40.9 & 67.6 & 56.7 \\
\midrule
\multirow{3}{*}{4} & GPTQ & $\checkmark$ & 16.05 & 57.2 & 51.2 & 39.9 & 63.4 & 52.9 \\
& Absmax & $\checkmark$ & 4.8e4 & 55.8 & 50.9 & 25.0 & 53.1 & 46.2 \\
& SmoothQuant & $\checkmark$ & 1.6e6 & 53.7 & 48.3 & 24.8 & 53.6 & 45.1 \\
\midrule
\multirow{2}{*}{2} & GPTQ & $\checkmark$ & 1032 & 51.6 & 50.1 & 25.8 & 53.4 & 45.2 \\
& QuIP & $\checkmark$ & 70.43 & 56.1 & 51.2 & 30.3 & 58.4 & 49.0 \\
\midrule
\multirow{2}{*}{1} & Absmax & $\checkmark$ & 3.5e23 & 49.8 & 50.0 & 24.8 & 53.6 & 44.6 \\
& SmoothQuant & $\checkmark$ & 3.3e21 & 50.5 & 49.5 & 24.6 & 53.1 & 44.4 \\
\midrule
\multirow{1}{*}{1}& BitNet & $\times$ & 17.07 & 66.3 & 51.4 & 38.9 & 66.9 & 55.9 \\
\bottomrule
\end{tabular}
\caption{Zero-shot results for BitNet and the baseline. PTQ, WGe, WG, SC, and HS are abbreviations for Post-training Quantization, Winogrande, Winograd, Storycloze, and Hellaswag datasets, respectively. When the weights are quantized to 1-bit, BitNet outperforms traditional PTQ methods. This table is taken from \cite{wang_bitnet}.}
\label{tab:Table 5}
\end{table}

\begin{table}[ht]
\centering
\renewcommand{\arraystretch}{0.8}
\small
\begin{tabular}{@{}lcccccccc@{}}
\toprule
\multirow{2}{*}{\textbf{Models}} &\multirow{2}{*}{ \textbf{Size}} &\multirow{2}{*}{ \textbf{WBits}} & \multicolumn{2}{c}{\textbf{7nm Energy (J)}} & \multicolumn{2}{c}{\textbf{45nm Energy (J)}} \\ 
\cmidrule(lr){4-5} \cmidrule(lr){6-7}
                &               &                & \textbf{MUL} & \textbf{ADD} & \textbf{MUL} & \textbf{ADD} \\ \midrule
\multirow{2}{*}{Transformer}     & \multirow{3}{*}{6.7B}          & 32             & 4.41         & 1.28         & 12.46        & 3.03         \\
                &               & 16             & 1.14         & 0.54         & 3.70         & 1.35         \\
BitNet          &               & 1              & \textbf{0.02} & \textbf{0.04} & \textbf{0.08} & \textbf{0.13} \\ \midrule
\multirow{2}{*}{Transformer}     & \multirow{3}{*}{13B}           & 32             & 8.58         & 2.49         & 24.23        & 5.89         \\
                &               & 16             & 2.23         & 1.05         & 7.20         & 2.62         \\
BitNet          &               & 1              & \textbf{0.04} & \textbf{0.06} & \textbf{0.12} & \textbf{0.24} \\ \midrule
\multirow{2}{*}{Transformer}     & \multirow{3}{*}{30B}           & 32             & 20.09        & 5.83         & 56.73        & 13.80        \\
                &               & 16             & 5.21         & 2.45         & 16.87        & 6.13         \\
BitNet          &               & 1              & \textbf{0.06} & \textbf{0.14} & \textbf{0.20} & \textbf{0.53} \\ 
\bottomrule
\end{tabular}
\caption{A comparison of energy consumption between BitNet and Transformer with an input length of 512. The bolded text represents the lowest energy consumption for that model size. Clearly, BitNet significantly reduces energy consumption. This table is sourced from \cite{wang_bitnet}.}
\label{tab:Table 6}
\end{table}

Additionally, they compared the energy consumption of BitNet and Transformer, as shown in Table \ref{tab:Table 6}. From Table \ref{tab:Table 6}, it is clear that for model sizes of 6.7B, 13B, and 30B, BitNet has the lowest energy consumption. This demonstrates that binarized LLMs significantly reduce computational costs, providing convenience for deployment.

\subsection{Comparison of Accuracy in Other Binarized LLM Techniques}

The BitNet a4.8 team \cite{wang_bitnet_a48} also compared the results of BitNet a4.8 \cite{wang_bitnet_a48}, BitNet b1.58 \cite{ma_era_1bit}, and LLaMA LLM on downstream tasks, as shown in Table \ref{tab:Table 7}.

\begin{table}[ht]
\centering
\renewcommand{\arraystretch}{0.8}
\small
\begin{tabular}{@{}llccccccc@{}}
\toprule
\textbf{Models}     & \textbf{Size} & \textbf{PPL$\downarrow$} & \textbf{ARCc$\uparrow$} & \textbf{ARCe$\uparrow$} & \textbf{HS$\uparrow$} & \textbf{PQ$\uparrow$} & \textbf{WGe$\uparrow$} & \textbf{Avg$\uparrow$} \\ \midrule
LLaMA LLM           & \multirow{4}{*}{700M} & 11.44         & 27.13          & 43.27          & 44.70        & 68.12        & 53.99        & 47.44        \\
BitNet b1.58        &                   & 12.32         & 25.00          & 42.68          & 42.08        & 66.97        & 54.14        & 46.17        \\
BitNet a4.8 (FP4)   &                   & 12.40         & 25.17          & 42.68          & 42.36        & 66.27        & 52.96        & 45.89        \\
BitNet a4.8         &                   & 12.40         & 25.17          & 41.58          & 42.44        & 66.38        & 53.04        & 45.72        \\ \midrule
LLaMA LLM           & \multirow{4}{*}{1.3B} & 10.82         & 27.90          & 45.16          & 47.65        & 69.91        & 53.35        & 48.79        \\
BitNet b1.58        &                   & 11.27         & 27.65          & 45.33          & 46.86        & 68.39        & 54.06        & 48.46        \\
BitNet a4.8 (FP4)   &                   & 11.38         & 28.50          & 44.36          & 47.03        & 68.61        & 54.06        & 48.51        \\
BitNet a4.8         &                   & 11.35         & 28.50          & 44.15          & 46.98        & 68.34        & 54.14        & 48.42        \\ \midrule
LLaMA LLM           & \multirow{4}{*}{3B}   & 9.61          & 29.95          & 48.11          & 55.25        & 71.76        & 57.46        & 52.51        \\
BitNet b1.58        &                   & 9.97          & 29.27          & 49.41          & 54.42        & 70.89        & 57.54        & 52.30        \\
BitNet a4.8 (FP4)   &                   & 9.99          & 29.10          & 49.24          & 54.60        & 71.38        & 56.12        & 52.08        \\
BitNet a4.8         &                   & 9.97          & 28.33          & 49.58          & 54.62        & 71.16        & 54.38        & 51.61       \\ \midrule
LLaMA LLM           & \multirow{4}{*}{7B}   & 9.20          & 33.36          & 51.22          & 58.33        & 73.34        & 58.41        & 54.93        \\
BitNet b1.58        &                   & 9.24          & 32.00          & 50.88          & 59.79        & 72.96        & 59.83        & 55.09        \\
BitNet a4.8 (FP4)   &                   & 9.42          & 31.57          & 51.22          & 58.20        & 72.47        & 59.59        & 54.61        \\
BitNet a4.8         &                   & 9.37          & 31.66          & 50.88          & 58.78        & 73.01        & 59.35        & 54.74        \\ 
\bottomrule
\end{tabular}
\caption{Performance comparison across models of different sizes, where the model sizes are 700M, 1.3B, 3B, and 7B. This table is provided by \cite{wang_bitnet_a48}.}
\label{tab:Table 7}
\end{table}

Table \ref{tab:Table 7} illustrates that as the model size increases, the performance gap between the full-precision (FP16) LLaMA large language model and BitNet b1.58 gradually narrows. Particularly for the 7B model, BitNet b1.58 shows comparable performance to LLaMA LLM in terms of language model perplexity and average accuracy on downstream tasks. BitNet a4.8 nearly achieves performance similar to BitNet b1.58.

\begin{table}[ht]
\centering
\renewcommand{\arraystretch}{0.8}
\small
\resizebox{\linewidth}{!}{%
\scriptsize
\setlength{\tabcolsep}{2pt}
\begin{tabular}{@{}l|lc|cccccccc|ccc@{}}
\hline
\multirow{2}{*}{\textbf{Model}} & \multirow{2}{*}{\textbf{Size}} & \multirow{2}{*}{\textbf{BW}} & \multicolumn{8}{c|}{\textbf{Zero-shot Accuracy} $\uparrow$} & \multicolumn{3}{c}{\textbf{Perplexity} $\downarrow$} \\

  &  &  & \textbf{BoolQ} & \textbf{PIQA} & \textbf{HS} & \textbf{WG} & \textbf{ARC-e} & \textbf{ARC-c} & \textbf{OBQA} & \textbf{Ave.} & \textbf{Wiki2} & \textbf{PTB} & \textbf{C4} \\
\hline
\textbf{BitNet b1.58} & 700M & 1.59 & 58.2 & \textbf{68.1} & \textbf{35.1} & \textbf{55.2} & \textbf{51.8} & \textbf{21.4} & 20.0 & \textbf{44.3} & \textbf{17.1} & \textbf{72.1} & \textbf{17.5} \\
FBI-LLM                 & 130M          & 1.01         & \textbf{62.1}          & 59.3          & 28.7          & 51.0          & 34.9          & 20.5          & \textbf{26.4}          & 40.4          & 28.2          & 136.6         & 26.9         \\ \midrule
TinyLLaMA            & 1.1B          & 16         & \uline{57.8}          & \uline{73.3}          & \uline{59.2}          & 59.1          & \uline{55.3}          & \uline{30.1}          & \uline{36.0}          & \uline{53.0}          & \uline{7.8}          & 30,5          & \uline{9.9}          \\
OPT                  & 1.3B          & 16         & \uline{57.8}          & 72.5          & 53.7          & \uline{59.5}          & 51.0          & 29.5          & 33.4          & 51.1          & 14.6          & \uline{20.3}          & 16.1          \\
OneBit-OPT            & 1.3B          & 1.02         & 59.5          & 62.6          & 34.3          & 51.1          & 41.3          & 24.1          & -          & -         & 25.4          & -         & 23.0          \\
BitNet b1.58          & 1.3B          & 1.59         & 56.7          & 68.8          & 37.7          & \textbf{55.8}          & \textbf{54.9}          & 24.2          & 19.6          & 45.4          & 24.1          & 145.1         & 21.8          \\
FBI-LLM                  & 1.3B          & 1.01         & \textbf{60.3}          & \textbf{69.0}          & \textbf{42.3}          & 54.0          & 43.6          & \textbf{25.3}          & \textbf{29.6}          & \textbf{46.3}          & \textbf{12.6}          & \textbf{39.3}          & \textbf{13.8}          \\ \midrule
OPT                  & 7B            & 16         & 66.1          & 76.5          & 67.2          & 65.4          & 60.0          & 34.7          & 37.4          & 58.2          & 10.9           & \uline{15.8}          & 127           \\
LLaMA                & 7B            & 16         & 75.1          & \uline{79.2}          & \uline{76.2}          & \uline{69.9}          & 72.9          & 44.9          & \uline{44.4}          & 66.0          & 5.7           & 41.2          & \uline{7.3}           \\
LLaMA2                & 7B            & 16       & \uline{77.7}          & 79.1         & 76.0          & 69.1          & \uline{74.6}          & \uline{46.2}          & 44.2          & \uline{66.7}          & \uline{5.5}           & 37.9          & \uline{7.3}           \\
OneBit-LLaMA2         & 7B            &          & \textbf{63.1}          & 68.1          & 52.6          & 58.4          & 41.6          & 29.6          & -          & -          & 9.7           & -          & 11.1           \\
BitNet         & 7B            & -         & -          & -          & 38.9          & 51.4          & -          & -          & -          & -          & -           & -          & -           \\

BiLLM-OPT            & 7B & 1.11 & 62.2 & 58.6 & 31.9 & 51.5 & 34.1 & 23.9 & 29.0 & 41.6 & 35.4 & 73.6 & 43.2 \\
                              BiLLM-LLaMA & 7B                & 1.08         & 62.7          & 61.2          & 36.8          & 51.1          & 36.0          & 25.7          & 31.8          & 43.6          & 35.0          & 421.3          & 39.6          \\

BiLLM-LLaMA2         & 7B                & 1.08         & 61.8          & 60.6          & 34.8          & 52.4          & 36.2          & 24.4          & 33.2          & 43.3          & 32.5           & 3877.4          & 40.5          \\ 
FBI-LLM             & 7B             & 1.01         & 61.5           & \textbf{72.6}                & \textbf{57.7}        & \textbf{58.9}           & \textbf{53.0}           & \textbf{29.9}           & \textbf{36.8}          & \textbf{52.9}            & \textbf{9.1}         & \textbf{29.6}        & \textbf{10.5}   \\
\bottomrule
\end{tabular}%
}
\caption{A comparison of Zero-shot Accuracy and Perplexity on downstream tasks across different networks. Here, BW refers to the average number of bits per parameter. Bold values indicate the best performance among non-high-precision models, while underlined values indicate the best performance among high-precision models. The model sizes range from 700M to 7B. These results are sourced from \cite{ma_fbi_llm}.}
\label{tab:Table 8}
\end{table}

In the FBI-LLM paper by Liqun Ma et al. \cite{ma_fbi_llm}, various methods are compared in terms of downstream task performance and perplexity, as shown in Table \ref{tab:Table 8}. From Table \ref{tab:Table 8}, it can be seen that FBI-LLM maintains the lowest average bit-width across different model sizes. Despite a fivefold difference in model size and quantization between FBI-LLM 130M and BitNet b1.58 700M, FBI-LLM outperforms BitNet b1.58 on BoolQA and OpenbookQA. For the 7B model size, FBI-LLM significantly outperforms nearly all baselines. However, we found that this improvement is largely due to the inclusion of distillation, and if distillation is removed, performance declines significantly, even resulting in non-convergence.
\subsection{Discussion}
We observed an interesting phenomenon. Referring to the methodology of PC++ \cite{lu2024understanding} in the BNN-CNN domain, we initially assumed that further smoothing of the gradient in BNN-LLMs would guarantee further convergence of the algorithm. However, in practice, we found that when using the Hard Tanh activation, due to the high curvature at the inflection point, even with a small learning rate, large gradients still appear during backpropagation, leading to non-convergence. On the other hand, using STE allows for convergence. Therefore, for BNN-LLMs, merely smoothing the gradient does not guarantee convergence. Additional conditions are required to ensure convergence, which calls for further research.

\begin{figure}[H]
    \centering
    \includegraphics[page=1, trim=10 10 10 10, clip, width=\textwidth]{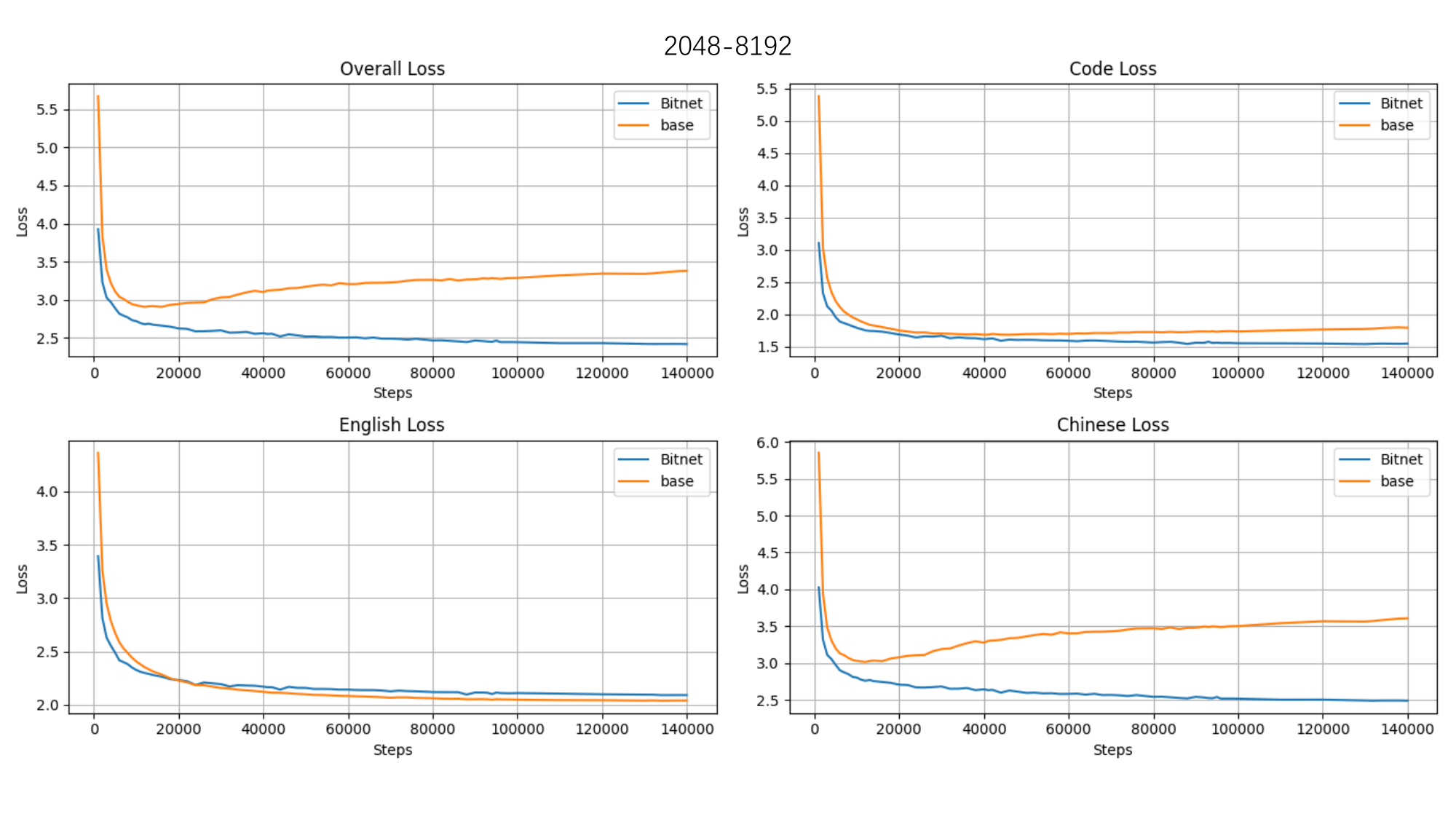} 
    \caption{The experimental results of BitNet 1.8B with input text lengths ranging from 2048 to 8192, where the yellow line/blue line represents the baseline/BitNet, respectively.} 
    \label{fig:Figure 7}  % 规范标签命名
\end{figure}

Furthermore, from-scratch quantization training in BNN-LLMs is regarded as a potential regularization technique in our view. This is because quantization implicitly introduces regularization by constraining the weight space, such as $\{-1, +1\}$ or $\{-1, 0, 1\}$, which is equivalent to adding a penalty term in the loss function.

\begin{equation}
\min_w \ell(w) + \lambda \cdot \|w - Q(w)\|_2^2
\end{equation}
Here, $Q$ is the quantization function, such as Sign, and $\lambda$ controls the strength of regularization. Our experiments further support this conclusion. 
We conducted experiments on BitNet \cite{wang_bitnet} 1.8B with input text sequences ranging from 2048 to 8192 in length. The results, shown in Figure \ref{fig:Figure 7}, reveal that the baseline model exhibits divergent loss behavior, while BitNet's loss continuously decreases. Therefore, it can be inferred that from-scratch quantization training may serve as a regularization technique.

\section{Conclusion and Future Trends}

\subsection{Summary of Binarization Application Areas}

The concept of binarized weights predates the rise of deep learning \cite{saad1990training}. However, early binarized networks only included a single hidden layer \cite{saad1990training,baldassi2007efficient}. In these early studies, weights could not be updated through small increments, making backpropagation ((BP) and stochastic gradient descent (SGD) unsuitable for these networks. Early binarization research mainly relied on variants of Bayesian inference. Later, Courbariaux proposed the BinaryConnect \cite{courbariaux_binaryconnect} method, which was applied to deep neural networks (DNNs). This led to the development of several algorithms, such as PC \cite{dockhorn2021demystifying}, BNN \cite{hubara2016binarized}, BNN+ \cite{darabi2019regularized}, BNN++ \cite{lu2024understanding}, and others. For large language models, Hongyu Wang et al. \cite{wang_bitnet} drew inspiration from BNN-CNN and proposed BitNet. Based on this, numerous algorithms were born, including BitNet b1.58 \cite{ma_era_1bit}, BitNet a4.8 \cite{wang_bitnet_a48}, FBI-LLM \cite{ma_fbi_llm}, Bi-Mamba \cite{tang_bi_mamba}, and others. Given the success of BitNet b1.58, Jainaveen Sundaram et al. \cite{sundaram2024llavaolmobitnet1b} attempted to apply the BitNet b1.58 idea to the multimodal domain. Specifically, they built LLaVaOLMoBitNet1B, a multimodal large language model based on the ternary OLMoBitNet1B \cite{olmobitnet1b} and LLaVa \cite{liu2024improved} methods. Compared to other multimodal large language models, they achieved the smallest number of parameters and the fewest pretraining tokens compared to full-precision counterparts, providing a baseline for future development of more powerful ternary multimodal models. Additionally, Jacob Nielsen et al. \cite{nielsen_bitnet_reloaded} proposed a variant of BitNet b1.58 called BitNet 1.58 Reloaded and applied it to small language models (SLMs) and vision models. They concluded that their method achieves near-state-of-the-art performance on small language models and surpasses state-of-the-art performance on vision models. Yiwei Lu et al. \cite{lu2024understanding} also applied this binarization technique to the Vision Transformer (VIT) domain, significantly reducing overhead. For example, they reduced the memory requirement of VIT-B from 450MB to 15MB, achieving approximately 30 times compression. For VIT, BNN++ \cite{lu2024understanding} shows significant advantages on most tasks.

\subsection{Future Research Trends}

In the subsequent research on BNN-LLM, it is anticipated that the future research trends will focus on the following: (1) Further reducing training costs, as we observed that BitNet b1.58 training is 60\% slower than the baseline and still relatively costly. (2) Further improvement of FBI-LLM. We found that for FBI-LLM, removing distillation leads to a significant performance drop and even non-convergence, so further optimization of FBI-LLM will be a new research direction. (3) Currently, there is no research on binarization or ternary quantization techniques in the Jamba domain. Therefore, applying binarization techniques to the Jamba architecture would be a valuable research direction. Additionally, it is worth considering applying 1-bit quantization techniques to MOE (Mixture of Experts). (4) For activation quantization, the MatMul-free method can be referenced to further reduce training costs. (5) Dynamically adjust the quantization strategy for weights and activation values based on input data and task requirements. (6) Design dedicated hardware that supports integer operations (rather than floating-point operations) to reduce energy consumption. (7) Optimize the memory architecture (such as SRAM) to reduce data transfer bottlenecks between the main memory and the chip. (8) Design specific operators that support low-bit models (e.g., binarized weight matrix operations: $\{-1, 1\}$).

\subsection{Conclusion}

In recent years, the technique of binarizing weights has seen extensive research and application in the field of deep learning. From deep neural networks to large language models, binarization has been widely adopted, with various technologies emerging, such as BNN-CNN and BNN-LLM. Many methods in BNN-CNN are capable of achieving significant model compression while maintaining accuracy close to that of full-precision models.

In the LLM domain, BitNet \cite{wang_bitnet} and BitNet b1.58 \cite{ma_era_1bit} have significantly reduced storage and computational complexity through weight binarization and ternarization. FBI-LLM \cite{ma_fbi_llm} successfully trained a fully binarized model through autoregressive distillation, demonstrating the feasibility of training from scratch. Meanwhile, Bi-Mamba \cite{tang_bi_mamba} combined binarization with state space models, showing high computational efficiency in long-sequence modeling tasks.

From the research of many scholars on binarization techniques, it is evident that as models continue to grow larger, applying binarization techniques to these large models is an effective strategy. Especially in the LLM domain, low-bit quantization models have become a key solution to addressing the high computational complexity and energy consumption bottlenecks of LLMs. These low-bit quantization models are expected to find even broader applications in the future, enabling the efficient deployment of large language models on mobile devices, edge computing, and cloud inference, thus providing robust technical support for the widespread adoption of artificial intelligence.

\printbibliography

\end{document}